\documentclass{article}

\usepackage{microtype}
\usepackage{graphicx}
\usepackage{subcaption}
\usepackage{booktabs}      
\usepackage{makecell}      
\usepackage{multirow}      
\usepackage{hyperref}

\usepackage[preprint]{icml2026}

\usepackage{amsmath}
\usepackage{amssymb}
\usepackage{mathtools}
\usepackage{amsthm}
\usepackage[capitalize,noabbrev]{cleveref}

\theoremstyle{plain}

\theoremstyle{definition}

\theoremstyle{remark}

\usepackage[textsize=tiny]{todonotes}

\icmltitlerunning{V-VLAPS: Value-Guided Planning for VLA Models}

\begin{document}

\twocolumn[
  \icmltitle{V-VLAPS: Value-Guided Planning for Vision-Language-Action Models}

  \icmlsetsymbol{equal}{*}
  \icmlsetsymbol{lead}{\textdagger}
  \begin{icmlauthorlist}
    \icmlauthor{Ke Ren}{equal,lead,ubc}
    \icmlauthor{Ali Salamatian}{equal,lead,ubc}
    \icmlauthor{Kieran Pattison}{equal,ubc}
    \icmlauthor{Cyrus Neary}{ubc}
  \end{icmlauthorlist}
  \icmlaffiliation{ubc}{The University of British Columbia}
  \icmlcorrespondingauthor{Ke Ren}{kren04@student.ubc.ca}

  \icmlkeywords{Vision-Language-Action models, Monte Carlo Tree Search, Value Functions, Robotics, Offline-to-Online}

  \vskip 0.3in
]

\printAffiliationsAndNotice{\icmlEqualContribution\textsuperscript{\textdagger}\,Project lead.}

\begin{abstract}

Vision-language-action (VLA) models provide strong action priors for robotic manipulation, but their reactive behavior can fail under distribution shift and long-horizon task structure. Recent VLA-guided planning methods improve execution by using pretrained policies to guide tree search, yet node selection still depends heavily on policy priors and visit-count exploration. Consequently, when the policy favors poor actions, the planner lacks a learned value signal to correct this bias. Prior work has shown that VLA representations encode rollout success and failure information, suggesting that they may also support value estimation during planning. We introduce Value-Guided Vision-Language-Action Planning and Search (V-VLAPS), which augments VLA-guided planning with a lightweight value head trained on offline VLA rollouts to predict Monte Carlo returns. These predictions guide Monte Carlo Tree Search in simulation toward higher-value branches. Across five LIBERO suites, V-VLAPS matches value-free planning baseline at the default search budget in aggregate, and analysis shows that many hard failures are root-level timeouts where predicted values are weakly separated. With a larger search budget, V-VLAPS improves over the baseline in all task suites with $+6$ percentage points on LIBERO-Object and $+4$ percentage points on LIBERO-10. Our results suggest that VLA representations can support not only failure prediction, but also value-guided planning when search reaches branches where value-based ranking matters.

\end{abstract}

\section{Introduction}
\label{sec:introduction}

Deploying robotic policies in open-world settings requires reliability under distribution shift. Recent advances in robot learning have been driven by large-scale Vision-Language-Action (VLA) models, transformer policies that predict action sequences from multimodal observations and language instructions ~\citep{rt1,rt2,octomodelteam2024octoopensourcegeneralistrobot, openVLA, pi_zero}. While VLA models serve as effective priors for generalist behaviors, they are fundamentally limited by their reliance on behavior cloning. As a result, they often exhibit brittle behavior when facing out-of-distribution (OOD) states.

A promising approach to address this problem is to augment the pretrained model with a planning search algorithm that evaluates possible futures in simulation. This requires a simulator or world model. In our experiments, this role is played by the LIBERO simulator. One such search algorithm is Monte Carlo Tree Search (MCTS), which builds a search tree by simulating candidate action sequences before selecting an action to execute \citep{_wiechowski_2022}. Following \citet{neary2025improvingpretrainedvisionlanguageactionpolicies}, we can leverage a VLA model as a policy prior to guide the MCTS, using a visit-count heuristic to manage exploration. However, we observe that relying solely on the VLA prior is insufficient for robust long-horizon planning. In the formulation proposed by
\citet{neary2025improvingpretrainedvisionlanguageactionpolicies}, the search lacks a value function, meaning it has no grounded estimate of the expected future return. Consequently, if the VLA prior is inaccurate, assigning high probability to suboptimal actions, the planner has no mechanism to correct this bias other than exhaustive count-based exploration. In other words, the search relies on the imitation prior rather than the underlying reward structure.

The signal the search lacks may already live in the VLA's own internal features. \citet{gu2025safemultitaskfailuredetection} show that a small probe trained on the latent features of a generalist VLA predicts task success or failure with strong cross-task generalization. If a binary success signal generalizes across tasks, the same features may also carry information useful for estimating continuous value. The natural use of such a signal in our setting is not post-hoc monitoring but active search guidance: feeding the prediction into the MCTS scoring rule to bias node selection. In this sense, our setting is an offline-to-online decision-making problem: offline rollouts of a frozen VLA are used to learn a value signal, and that signal is then used at test time to guide online MCTS planning.

We introduce V-VLAPS, which extends the VLA-feature-probe approach of \citet{gu2025safemultitaskfailuredetection} from passive failure detection to active value-guided planning. A three-layer MLP trained on Octo readouts predicts Monte Carlo discounted returns, and we add the prediction as the $Q$ term in VLAPS's PUCT scoring rule. We evaluate V-VLAPS on five LIBERO suites against the plain Octo VLA and against VLAPS without a learned value, at two per-episode search budgets ($600$~s and $1800$~s). On these suites both MCTS-based conditions outperform the reactive Octo VLA by about $27$ percentage points on average, reproducing the result of \citet{neary2025improvingpretrainedvisionlanguageactionpolicies}; the value head is therefore evaluated as an addition to an already-strong planning baseline.

At the $600$~s budget, V-VLAPS matches VLAPS in aggregate (both at $87.4\%$), and a failure-mode analysis shows that many hard failures are MCTS timeouts at the root rather than clear value-head errors. Near the root, candidate branches are still far from task completion, so the value head often assigns similarly small values to many alternatives and has little discriminative power. With the budget extended to $1800$~s, search can reach deeper states where branch values separate more clearly. In this regime, V-VLAPS exceeds VLAPS by $+6$pp on libero\_object and $+4$pp on libero\_10. These results suggest that small probes trained on frozen VLA features can serve not only as failure detectors \citep{gu2025safemultitaskfailuredetection}, but also as learned planning components when search reaches states where value-based branch ranking is important.

\IfFileExists{VVLAPS.png}{%
\begin{figure*}[t]
    \centering
    \includegraphics[width=0.95\textwidth]{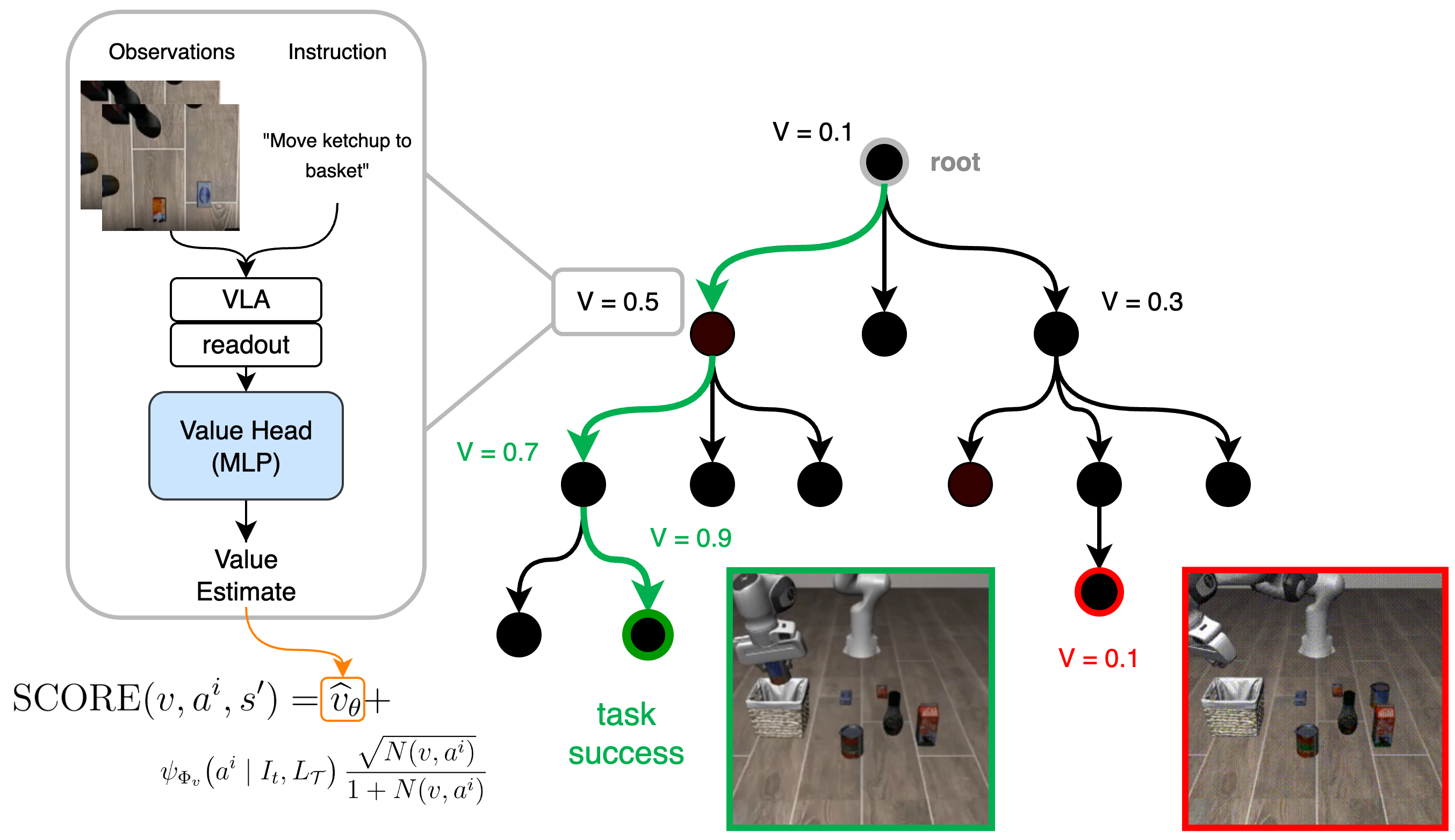}
    \caption{Overview of our value-guided VLAPS extension. At each MCTS node, the current observations and language instruction (e.g., ``Move ketchup to basket'') are passed through the frozen VLA backbone and our value head (MLP) to produce a scalar value estimate. This value is attached to the corresponding node and used in the VLAPS scoring rule to bias node selection. Nodes with higher predicted value (green) are selected more often and tend to lead to successful task completions, while low-value nodes (red) are down-weighted during search.}
    \label{fig:overall}
\end{figure*}
}{}

\section{Related Work}
\label{sec:related_work}

\subsection{Vision–Language–Action Policies}
Vision--language--action (VLA) models are designed to map visual observations and natural-language instructions directly to robot actions. Early systems showed that a single transformer policy can be trained on large collections of robot demonstrations and language-labelled tasks, and then be reused across many manipulation skills~\citep{rt1,rt2}. In our work, we use Octo~\citep{octomodelteam2024octoopensourcegeneralistrobot} as a fixed generalist VLA backbone and build our method on top of its latent representation; the same method can be used with other VLA models.

Although VLA models can perform a wide range of tasks, they are typically used in a purely reactive way: at each step, the model receives the current observation and instruction and outputs the next action (or action chunk), without explicit long-horizon planning.
In challenging scenes or out-of-distribution configurations, this can lead to failures that the policy does not recover from \citep{gu2025safemultitaskfailuredetection}. In the language modeling paradigm, many methods have been introduced to mitigate similar issues by scaling test-time compute, such as chain-of-thought prompting \citep{wei2023chainofthoughtpromptingelicitsreasoning}, self-consistency sampling \citep{wang2023selfconsistencyimproveschainthought}, and MCTS during inference \citep{hao2023reasoninglanguagemodelplanning}.

Recent work has also begun to study how extra test-time computation can improve VLA execution rather than relying solely on one-shot reactive action prediction \citep{robomonkey, TACO}. \citet{neary2025improvingpretrainedvisionlanguageactionpolicies} introduce Vision-Language-Action Planning and Search (VLAPS), which embeds a pre-trained VLA into a model-based search procedure and uses the VLA to define action proposals and abstractions for planning. In our work, we build on VLAPS and keep the same fixed VLA backbone, but additionally learn a value function over its latent state and incorporate the resulting value predictions into VLAPS's scoring rule to guide search.

\subsection{Monte Carlo Tree Search}

A natural way to compensate for the myopic behaviour of reactive policies is to add a search procedure that can simulate possible future outcomes before committing to an action. Monte Carlo Tree Search (MCTS) is a widely used algorithm for this purpose. It incrementally builds a search tree by repeatedly simulating trajectories from the current state, and at each iteration it selects actions that balance exploring uncertain branches with exploiting branches that have produced good returns in previous simulations~\citep{_wiechowski_2022}. Typically, each iteration proceeds through four phases: selection, expansion, simulation, and backpropagation. The resulting statistics over node visit counts and returns are then used to estimate action values at the root.

MCTS has been particularly successful in domains where a reasonably accurate simulator is available. In board games such as Go, Chess, and Shogi, MCTS combined with learned policies and values has led to systems that reach and surpass human expert performance~\citep{AlphaGo,silver2017masteringchessshogiselfplay}. In these settings, the search runs entirely in simulation and only the final chosen move is executed in the real environment. \citet{neary2025improvingpretrainedvisionlanguageactionpolicies} bring this style of planning to vision--language--action models in the VLAPS framework. They use MCTS over discrete action chunks, with a pre-trained VLA policy providing action priors that focus and refine the search of an otherwise completely intractable space. Their formulation combines their VLA-informed prior with visit-count-based exploration heuristics in a PUCT-style scoring rule which allows MCTS to exploit the strengths of the pre-trained policy while still exploring alternative action sequences.

\subsection{Learned Value Functions for Tree Search and VLA Planning}

Combining learned value functions with tree search has proved highly effective in domains where planning is possible. In AlphaGo and its successors, deep neural networks predict both a policy over moves and a scalar value estimate from a board position; Monte Carlo Tree Search uses the policy to prioritize promising actions and the value to evaluate leaf nodes without long rollouts~\citep{AlphaGo,silver2017mastering,silver2017masteringchessshogiselfplay}. This combination allows the search procedure to concentrate computation on moves that are both likely under the learned policy and lead to positions that the value function predicts as strong, rather than relying solely on visit counts and simulation outcomes of intractably long rollouts.

VLAPS adopts a similar strategy for vision--language--action policies, but stops short of learning an explicit value function over the VLA latent state. In \citet{neary2025improvingpretrainedvisionlanguageactionpolicies}, the quality of a node is determined by search statistics and VLA-derived action priors, but there is no separate learned estimate of the state's value. In contrast, we keep the underlying VLA and search procedure fixed, and introduce a small value head which takes the VLA latent state as input. This head is trained to predict Monte Carlo returns from VLA rollouts, and its predictions are incorporated into the VLAPS selection rule. Conceptually, this brings the use of value functions in VLAPS closer to the value-guided tree search used in game-playing systems like AlphaGo, while operating in the setting of vision--language--action control.

More directly related is SAFE \citep{gu2025safemultitaskfailuredetection}, which trains a small probe (an MLP or LSTM) on the VLA's frozen latent features and uses the resulting success/failure prediction as a post-hoc monitor. Our value head adopts the same feature source and probe family but predicts a continuous Monte Carlo discounted return, and the prediction enters the PUCT scoring rule during search rather than after the episode.

\section{Method}
\label{sec:methodology}

We extend VLAPS by learning a value function over Octo latent states and using this value to guide Monte Carlo Tree Search in simulation. As in VLAPS, the search uses a simulator to roll candidate action chunks forward before execution. We first collect rollouts of the Octo model and compute Monte Carlo value targets for each state (Section~\ref{sub:data_collection}). We then train a three-layer MLP (the ``value head'') to predict these targets from Octo's last-layer representation (readout) (Section~\ref{sub:value_training}). Finally, we integrate the learned value into VLAPS’ tree search by modifying the selection score to prefer action chunks that lead to high-value successor states (Section~\ref{sub:value_integration}).

\subsection{Data Collection}
\label{sub:data_collection}
We construct training data for the value head by rolling out a fixed pre-trained VLA policy, without any planning, on LIBERO~\citep{liu2023libero} tabletop manipulation tasks. For each suite, we collect value-head training rollouts on all tasks but restrict each task to four of the ten available initial states: indices $0$, $3$, $6$, and $9$. Evaluation uses all ten initial states per task (Section~\ref{sec:experiments}), so six initial states per task are unseen during value-head data collection. This gives a partial initial-state generalization test while preserving comparability with the full ten-initial-state evaluation used for all methods.

At the beginning of each episode, the environment provides an initial observation $o_0$ and a language instruction $g$. At each decision step $t$, the VLA (Octo~\citep{octomodelteam2024octoopensourcegeneralistrobot}) receives $(o_t, g)$, computes a latent readout vector $h_t \in \mathbb{R}^d$ which acts as the summary representation of the past observations and the task. It then uses the readout to output an action chunk $c_t$, which consists of a sequence of low-level actions. We execute the entire chunk $c_t$ in the environment, applying each low-level action in order, until the chunk finishes or the episode terminates. The environment then returns the next observation $o_{t+1}$, and we query the VLA again. This produces an episode as a sequence of decision steps $(o_0, c_0, o_1, c_1, \dots, o_T)$.

Episodes terminate either when the task is successfully completed or when a timeout horizon is reached. We use a sparse terminal value: if the task is successfully completed before the timeout, the episode receives a terminal value of $1$; otherwise (failure or timeout), it receives a terminal value of $0$.

For each decision step $t$ in an episode, we define a Monte Carlo value target by propagating the terminal value backward through time. Let $R \in \{0, 1\}$ denote the terminal value and $T$ the index of the final decision step. We assign each state at step $t$ a target
\[
G_t =
\begin{cases}
\gamma^{T - t}, & \text{if the episode ends in success } (R = 1), \\
0,              & \text{if the episode ends in failure } (R = 0).
\end{cases}
\]
where $\gamma \in (0, 1]$ is a discount factor; we use $\gamma = 0.99$. This target is a Monte Carlo estimate of the state value under our sparse success value when following the VLA policy.

We pair each readout with its value target to form a training example $(h_t, G_t)$, and aggregate these across many rollouts to build a per-suite dataset of state--value pairs. The raw target distribution is heavily skewed toward zero: failed episodes contribute only zero-valued targets, and even successful episodes have many early-step targets close to zero because $\gamma^{T-t}$ is small when $T-t$ is large. A value head trained on this distribution can collapse to predicting zero almost everywhere, so we rebalance once during preprocessing: targets are bucketed into ten equal-width bins on $[0,1]$, and the bottom bin is downsampled to match the combined size of the other nine bins.

\subsection{Value Head Training}
\label{sub:value_training}
Our goal is to learn a value function that maps the VLA's latent representation at each decision step to the scalar value target defined in Section~\ref{sub:data_collection}. Concretely, given the Octo readout vector $h_t \in \mathbb{R}^d$ at decision step $t$, we learn a function $V_\theta : \mathbb{R}^d \to \mathbb{R}$ that predicts a scalar estimate $\hat{v}_t$ of the state value: $\hat{v}_t = V_\theta(h_t)$.

We parameterize $V_\theta$ as a lightweight three-layer multilayer perceptron (MLP) with approximately $2.4$M parameters in total. The architecture is intentionally small compared to the underlying Octo model, so that evaluating the value head adds minimal overhead during planning.

We train this value head on the dataset of readout–value pairs $\{(h_t, G_t)\}$. The training objective is to regress onto the Monte Carlo value targets $G_t$ using mean squared error: $\mathcal{L}(\theta) = \mathbb{E}_{(h_t, G_t)} \bigl[(V_\theta(h_t) - G_t)^2\bigr]$.

In practice, we approximate this expectation by sampling mini-batches from the dataset and performing stochastic gradient descent with the Adam optimizer. During training, the parameters of the underlying Octo VLA are kept fixed; only the value head parameters $\theta$ are updated. After convergence, $V_\theta$ provides a compact estimate of the state value at each decision step, which we use to guide the tree search.

\subsection{Value-Guided Planning}
\label{sub:value_integration}

At each decision point, VLAPS builds a temporary search tree in simulation. The root node represents the current environment state before the next action chunk is chosen. Each edge corresponds to a candidate action chunk, and each child node represents the simulated state reached after executing that chunk. The search expands this tree until a simulated success is found or the wall-time budget is reached, then executes an action chunk from the root and replans from the next observation.

Motivated by the value-plus-prior structure of PUCT-style search, we integrate the learned value head into the VLAPS tree-search rule while leaving the rest of the VLAPS search procedure unchanged (Figure \ref{fig:overall}). VLAPS already provides the prior/exploration term through its VLA-guided action prior and visit statistics; V-VLAPS adds a learned value term for the successor state reached by each candidate chunk. At a node $v$, VLAPS samples a finite set of candidate action chunks ${a^i}$ from the VLA-induced action distribution and scores them using the VLAPS prior/exploration term

\begin{align}\label{eq:vlaps_score}
U_{\mathrm{VLAPS}}(v, a^i)
= \psi_{\Phi_v}\!\left(a^i \mid I_t, L_\mathcal{T}\right)
\frac{\sqrt{N(v, a^i)}}{1 + N(v, a^i)} ,
\end{align}
where $I_t$ is the current observation, $L_\mathcal{T}$ is the task instruction, $\psi_{\Phi_v}$ is the VLAPS prior over the sampled candidate chunks, and $N(v,a^i)$ is the number of times candidate $a^i$ has been selected from node $v$. We use this term exactly as in VLAPS.

V-VLAPS adds a learned value for the child node reached by each candidate chunk. For a candidate action chunk $a^i$ at node $v$, the simulator rolls the chunk forward to obtain the successor state $s’$. We compute the Octo readout $h(s’)$ for this state and evaluate the value head:
\begin{align}
    \widehat v_\theta(s') = V_\theta(h(s')) .
\end{align}
We use this successor-state value as the value term for the node-action pair,
\begin{align}
    Q_\theta(v,a^i) = \widehat v_\theta(s') .
\end{align}
Because the simulated transition for a fixed action chunk is deterministic in our LIBERO setup, the successor-state value estimates the continuation value after choosing $a^i$.

The final V-VLAPS selection score is
\begin{equation}\label{eq:vvlaps_score}
\begin{split}
    \mathrm{SCORE}(v, a^i)
    &=
    \lambda_V Q_\theta(v,a^i)
    +
    U_{\mathrm{VLAPS}}(v,a^i).
\end{split}
\end{equation}
We use a fixed value coefficient $\lambda_V = 1$ in all experiments.

Therefore, V-VLAPS differs from VLAPS only by adding the learned value term $Q_\theta$ with a fixed coefficient. The action sampler, action library, LIBERO simulator, and MCTS hyperparameters are kept fixed across the VLAPS and V-VLAPS conditions.

\section{Experiments}
\label{sec:experiments}

All experiments are conducted in LIBERO simulation. For VLAPS and V-VLAPS, the same simulator is used both for planning rollouts and for evaluation. This gives a controlled setting where the planner has access to an accurate model of the environment. We evaluate three methods on five LIBERO suites: the plain Octo VLA without planning (\textit{VLA}), VLAPS without a learned value (\textit{VLAPS}, the formulation of \citet{neary2025improvingpretrainedvisionlanguageactionpolicies}), and V-VLAPS, which adds the learned value head from \Cref{sub:value_integration} to the VLAPS scoring rule. The five suites cover different task types: \textit{libero\_object} tests object identification with pick-and-place, \textit{libero\_spatial} tests spatial reasoning between objects and target regions, \textit{libero\_goal} tests goal-conditioned behaviour, \textit{libero\_10} is ten long-horizon tasks, and \textit{libero\_90} is ninety diverse manipulation tasks~\citep{liu2023libero}. The value head is trained once from offline VLA rollouts for each suite and then frozen during evaluation. All test-time adaptation comes from MCTS planning rather than parameter updates.

Each (condition, suite, budget) cell is evaluated on $100$ episodes. For the four 10-task suites this is the full $10 \times 10$ grid of tasks and initial states, and for libero\_90 we sample $10$ tasks and use all $10$ initial states per sampled task. As described in \Cref{sub:data_collection}, the value head is trained using rollouts from only four initial states per task. The reported evaluation therefore tests both performance on the value-head data-collection states and generalization to six initial states per task that are unseen during value-head training. MCTS uses two per-episode wall-time budgets, $600$~s by default and $1800$~s for the extended-budget experiments. The MCTS hyperparameters are held fixed across all conditions and suites; the full configuration is in \Cref{app:hyperparameters}.

We train one value head per suite, with a $90/10$ train/validation split (fixed seed). Per-suite balanced training-data sizes range from $7{,}830$ examples (libero\_object) to $18{,}792$ (libero\_spatial); the full breakdown is in \Cref{app:dataset}.

\section{Results}
\label{sec:results}




We separate the benefit of planning from the additional benefit of value guidance. We first compare aggregate success rates at two wall-time search budgets, then inspect failures at the default budget to understand when the learned value term can and cannot affect search.

\subsection{Main results}
\label{sub:main_results}

The main pattern is that value guidance does not change aggregate performance at the default search budget, but improves success rate when the search budget is increased. Table~\ref{tab:main} reports success rates for the three methods on each of the five suites. At the $600$~s wall-time search budget, V-VLAPS and VLAPS both average $87.4\%$ across the five suites. No per-suite difference exceeds $3$pp, which sits inside the binomial standard error of $\approx 3$pp at $p \approx 0.9$. At the $1800$~s wall-time search budget, the gap widens to $+3.2$pp ($91.6\%$ for V-VLAPS vs.\ $88.4\%$ for VLAPS). The largest gains are on libero\_object ($+6$pp) and libero\_10 ($+4$pp). The other three suites already reach at least $88\%$ success at the $600$~s budget for both planning methods, and move by at most $\pm 2$pp at the larger search budget. Because each cell contains $100$ episodes, we interpret $1$--$3$pp differences as suggestive rather than conclusive and focus our claims on the larger gaps under the extended search budget.

\begin{table*}[t]
    \centering
    \caption{Success rate (\%) by method and suite at the default ($600$~s) and extended ($1800$~s) per-episode search budgets. Each cell aggregates $100$ episodes. \textbf{VLA} (no planning) provides context for the regime; \textbf{VLAPS} \citep{neary2025improvingpretrainedvisionlanguageactionpolicies} is the prior-work planning baseline; \textbf{V-VLAPS} is our extension. Bold V-VLAPS cells highlight the $1800$~s wall-time gains on the headroom suites (libero\_object, libero\_10).}
    \label{tab:main}
    \footnotesize
    \setlength{\tabcolsep}{6pt}
    \begin{tabular}{lrrrrr}
        \toprule
         &  & \multicolumn{2}{c}{$600$~s budget} & \multicolumn{2}{c}{$1800$~s budget} \\
        \cmidrule(lr){3-4} \cmidrule(lr){5-6}
        \textbf{Suite} & \textbf{VLA} & \textbf{VLAPS} & \textbf{V-VLAPS} & \textbf{VLAPS} & \textbf{V-VLAPS} \\
        \midrule
        libero\_object  & 37 & 82 & 85 & 87 & \textbf{93} \\
        libero\_spatial & 81 & 96 & 95 & 96 & \textbf{97} \\
        libero\_goal    & 88 & 92 & \textbf{93} & 90 & \textbf{93} \\
        libero\_10      & 38 & 77 & 75 & 81 & \textbf{85} \\
        libero\_90      & 57 & \textbf{90} & 89 & 88 & \textbf{90} \\
        \midrule
        \textbf{Avg.}   & 60.2 & 87.4 & 87.4 & 88.4 & \textbf{91.6} \\
        \bottomrule
    \end{tabular}
\end{table*}

\subsection{Wall-time scaling and a task-7 case study}
\label{sub:walltime}




On the two suites with the largest remaining headroom, increasing the wall-time search budget improves task success more for V-VLAPS than for VLAPS. On libero\_object, V-VLAPS increases from $85\%$ at $600$~s to $93\%$ at $1800$~s ($+8$pp), while VLAPS increases from $82\%$ to $87\%$ ($+5$pp). The V-VLAPS and VLAPS gap therefore grows from $+3$pp to $+6$pp. On libero\_10, the asymmetry is larger: V-VLAPS increases by $+10$pp ($75\% \to 85\%$), while VLAPS increases by $+4$pp ($77\% \to 81\%$), so the per-suite gap changes from $-2$pp to $+4$pp.

This difference is sharpest on a single hard task. To tighten the comparison on libero\_object task $7$, we extended the default ten initial states to twenty (indices $0$–$19$) at the $1800$s search budget. V-VLAPS reaches $17/20$ ($85\%$), while VLAPS reaches $13/20$ ($65\%$). The $N = 10$ data (Appendix \ref{app:pertask}) show the same pattern from the original evaluation set: when moving from $600$~s to $1800$~s, VLAPS goes from $8/10$ to $6/10$, while V-VLAPS goes from $5/10$ to $10/10$. On this task, extra search time helps the value-guided variant more than VLAPS without the learned value term.

libero\_spatial, libero\_goal, and libero\_90 are already at or above $88\%$ success at $600$~s, and neither the larger search budget nor the value head moves them by more than $\pm 2$pp.\footnote{VLAPS drops slightly on libero\_goal ($92\% \to 90\%$) and libero\_90 ($90\% \to 88\%$) at $1800$~s. Both moves are within sampling noise.}

\subsection{Failure-mode analysis}
\label{sub:failure_mode}


Root-level timeouts account for most failures on libero\_object tasks 6, 7, and 8 at the 600s search budget (Table \ref{tab:failure_modes}). We define a root-level timeout as a failed episode in which MCTS spends the full wall-time budget at the root node, before executing the first action chunk. These failures indicate that the search often does not reach deeper states where value-based ranking could change later decisions.

This pattern appears for both planning methods. In V-VLAPS, $13$ of the $15$ failed episodes are root-level timeouts. In VLAPS, $14$ of the $18$ failed episodes show the same behaviour. For V-VLAPS, these runs contain only the initial value-head evaluations at the root, with wall-clock times between $605$ and $611$~s when the search stops. The small overshoot past $600$~s comes from checking the time budget after a search iteration completes.

\begin{table}[H]
    \centering
    \caption{Failure modes on libero\_object tasks $6$–$8$ at the $600$~s search budget. A root-level timeout is a failed episode in which MCTS remains at the root node until the wall-time cap is reached, before executing the first action chunk.}
    \label{tab:failure_modes}
    \small
    \setlength{\tabcolsep}{7pt}
    \begin{tabular}{lrrr}
        \toprule
        \textbf{Method} & \textbf{Failures} & \makecell{\textbf{Root}\\\textbf{timeouts}} & \textbf{Fraction} \\
        \midrule
        VLAPS & 18 & 14 & 77.8\% \\
        V-VLAPS & 15 & 13 & 86.7\% \\
        \bottomrule
    \end{tabular}
\end{table}


For the V-VLAPS root-level timeouts, the value head predicts successor-state values between $-0.01$ and $0.02$. This narrow range is expected near the root of hard task configurations. Most candidate branches are still far from task completion, so their predicted continuation values are close to zero and weakly separated. In these near-root states, the value term has little ability to distinguish among alternatives. The extended-budget results are consistent with this interpretation: value guidance becomes more useful when MCTS reaches deeper states where different branches have more separated continuation values.

\section{Discussion and Conclusion}
\label{sec:discussion_and_conclusion}

V-VLAPS gains over VLAPS in a specific regime: when MCTS has time to move beyond near-root states, and when the task has remaining headroom for value-guided ranking to matter. Near the root of hard tasks, candidate branches often remain far from completion and receive similarly small value estimates, so the value term provides little separation. The first condition fails at $600$~s on the hardest libero\_object tasks, where most failures are root-level timeouts (\Cref{sub:failure_mode}). The second fails on libero\_spatial, libero\_goal, and libero\_90, where the prior pipeline is already near ceiling. The libero\_object and libero\_10 gains at $1800$~s are the regime where both conditions hold: search reaches deeper branches, and those branches have enough performance headroom for the value head's ranking to change decisions.

V-VLAPS shows that the VLA-feature probe \citet{gu2025safemultitaskfailuredetection} introduced for failure detection extends to active search guidance. With the binary failure target replaced by a Monte Carlo return and the prediction fed into the PUCT scoring rule, the same probe architecture can bias an MCTS search toward higher-return branches rather than only flag failures after the fact. We have demonstrated this for value-guided node ranking; whether the same probe family can drive other components of the planning loop, such as termination or branch pruning, is an open empirical question.

\paragraph{Limitations.} V-VLAPS requires a simulator or world model for search. In this paper, we evaluate in LIBERO, where the simulator used for planning matches the evaluation environment. Applying the same idea outside simulation would require an accurate learned or engineered world model. Our experiments use Octo as the VLA backbone, and evaluating whether the same pattern holds for other VLA models such as $\pi_0$~\citep{pi_zero} is an important next step. The value head is trained on rollouts of the unguided VLA, so its predictions on states the search visits are off-policy with respect to V-VLAPS execution. Per-suite training sets are small, ranging from $7{,}830$ to $18{,}792$ examples (Appendix~\ref{app:dataset}), and whether more training data would widen the $1800$~s gap or close the $600$~s gap is untested. In addition, a controlled ablation with shuffled value targets or a randomly initialized value head would further isolate the contribution of learned value estimates, and we leave this for future work.

\paragraph{Future work.} Two directions follow directly from the findings above. Training the value head on V-VLAPS rollouts instead of unguided VLA rollouts would address the off-policy gap, and is a prerequisite for iterative refinement of the value estimate. Using the value head to prune branches during expansion would convert the root-level timeouts of \Cref{sub:failure_mode} into search compute spent on the branches the value head ranks highest. Two further directions are more speculative. Replacing the MSE regression objective with the HL-Gauss classification head of \citet{farebrother2024stopregressing} is a natural alternative; we attempted it and hit class-imbalance issues that we expect would ease with more per-suite training data. Aggregating the value-head input across multiple transformer layers, as \citet{gu2025safemultitaskfailuredetection} suggest for SAFE, may expose signal the final readout averages out.

\section*{Impact Statement}

This paper studies value-guided MCTS planning over a fixed pretrained vision-language-action policy, evaluated in LIBERO simulation only. Our per-episode test-time search budgets are in the hundreds to thousands of seconds, which shifts compute from training to deployment with the associated energy cost. We do not foresee impacts specific to this contribution that are not already raised by the underlying VLA, planning, and failure-detection work this paper builds on.

\bibliographystyle{icml2026}
\bibliography{references}

\newpage
\appendix
\onecolumn

\section{Per-suite training-data sizes}
\label{app:dataset}

Table~\ref{tab:dataset} reports the per-suite value-head training-data sizes after rebalancing (\Cref{sub:data_collection}). These datasets are collected only from initial states $0$, $3$, $6$, and $9$ for each task, and the main evaluation in Table~\ref{tab:main} uses all ten initial states per task, including six initial states unseen during value-head training. For libero\_10, the raw distribution was already balanced ($50.0$\% zero), so its Balanced count equals its Raw count.

\begin{table}[h]
    \centering
    \caption{Per-suite training-data sizes after rebalancing. \textbf{Raw} is the count collected from VLA rollouts before rebalancing. \textbf{\% zero} is the fraction of raw examples with $G_t = 0$. \textbf{Balanced} is the count after downsampling the bottom bin to match the combined size of the other nine bins. \textbf{Train}/\textbf{Val} are the $90/10$ split sizes.}
    \label{tab:dataset}
    \footnotesize
    \setlength{\tabcolsep}{8pt}
    \begin{tabular}{lrrrrr}
        \toprule
        \textbf{Suite} & \textbf{Raw} & \textbf{\% zero} & \textbf{Balanced} & \textbf{Train} & \textbf{Val} \\
        \midrule
        libero\_object  & 60{,}695 & 92.8 & 8{,}700  & 7{,}830  & 870 \\
        libero\_spatial & 28{,}580 & 63.5 & 20{,}880 & 18{,}792 & 2{,}088 \\
        libero\_goal    & 34{,}155 & 70.3 & 20{,}300 & 18{,}270 & 2{,}030 \\
        libero\_10      & 11{,}020 & 50.0 & 11{,}020 & 9{,}918  & 1{,}102 \\
        libero\_90      & 63{,}595 & 86.8 & 16{,}810 & 15{,}129 & 1{,}681 \\
        \bottomrule
    \end{tabular}
\end{table}

\section{Value-head training and MCTS hyperparameters}
\label{app:hyperparameters}

This appendix lists the hyperparameters used to train the value head (Table~\ref{tab:vh-hyper}) and to run MCTS in the experiments of \Cref{sec:experiments} (Table~\ref{tab:mcts-hyper}). All evaluations were run on a single NVIDIA H100 GPU; one $100$-episode \textit{(suite, condition)} cell at the $600$~s budget completes in roughly $3.5$--$4.5$~h.

\begin{table}[h]
    \centering
    \caption{Value-head training hyperparameters (Section~\ref{sub:value_training}).}
    \label{tab:vh-hyper}
    \footnotesize
    \setlength{\tabcolsep}{8pt}
    \begin{tabular}{ll}
        \toprule
        \textbf{Setting} & \textbf{Value} \\
        \midrule
        Architecture (MLP) & $768 \to 3072$ (ReLU) $\to$ scalar \\
        Parameter count & $\sim 2.4$M \\
        Optimizer & Adam \\
        Initial learning rate & $1 \times 10^{-4}$ \\
        Learning-rate schedule & cosine decay ($\alpha = 0.1$) \\
        Batch size & $64$ \\
        Epochs & $100$ \\
        Train/val split & $90/10$ (fixed seed $42$) \\
        Loss & MSE on Monte Carlo returns $G_t$ \\
        Discount $\gamma$ & $0.99$ \\
        Target rebalancing & 10-bin equal-width on $[0,1]$, bottom bin downsampled \\
        \bottomrule
    \end{tabular}
\end{table}

\begin{table}[h]
    \centering
    \caption{MCTS hyperparameters, shared between VLAPS and V-VLAPS (Section~\ref{sub:value_integration}). The value head adds the $Q$ term in the SCORE function for the V-VLAPS condition only.}
    \label{tab:mcts-hyper}
    \footnotesize
    \setlength{\tabcolsep}{8pt}
    \begin{tabular}{ll}
        \toprule
        \textbf{Setting} & \textbf{Value} \\
        \midrule
        PUCT exploration weight $\psi$ & $1.0$ \\
        Action samples per node & $300$ \\
        Expansions per node & $10$ \\
        Expansions per MCTS step & $1$ \\
        Maximum tree depth & $80$ \\
        Maximum simulated rollout length & $300$ \\
        Chunk steps simulated per expansion & $4$ \\
        Replan after each executed chunk & yes \\
        Per-episode wall-time cap & $600$~s (default budget), $1800$~s (extended budget) \\
        Action chunk library size & $2{,}000$ medoids of all-suite VLA successes \\
        $\beta_\Phi$ distribution (action sampler) & softmax over chunk distances, $\alpha = 10$, $\epsilon = 0.1$ \\
        $\psi_{\Phi_v}$ distribution (PUCT prior) & softmax over sampled chunks, $\alpha = 5$ \\
        VLA generation temperature & $0.0$ \\
        Terminate search on simulated success & yes \\
        \bottomrule
    \end{tabular}
\end{table}

\section{Per-task results on libero\_object}
\label{app:pertask}

Table~\ref{tab:pertask-object} decomposes the libero\_object row of Table~\ref{tab:main} by task. Tasks $0$--$5$ and $9$ sit at or near the $90$--$100\%$ ceiling for both conditions at both budgets; the variance that drives the suite-level differences is concentrated on tasks $6$, $7$, and $8$. The $N = 20$ task-$7$ result of \Cref{sub:walltime} is a tighter measurement of the same task.

\begin{table}[h]
    \centering
    \caption{Per-task success counts on libero\_object at both per-episode wall-time budgets, $10$ initial states per task. Full-suite totals match the libero\_object row of Table~\ref{tab:main}. The bold cell highlights the V-VLAPS $5/10 \to 10/10$ swing on task $7$ at the larger budget.}
    \label{tab:pertask-object}
    \footnotesize
    \setlength{\tabcolsep}{12pt}
    \begin{tabular}{rrrrr}
        \toprule
         & \multicolumn{2}{c}{$600$~s budget} & \multicolumn{2}{c}{$1800$~s budget} \\
        \cmidrule(lr){2-3} \cmidrule(lr){4-5}
        \textbf{Task} & \textbf{VLAPS} & \textbf{V-VLAPS} & \textbf{VLAPS} & \textbf{V-VLAPS} \\
        \midrule
        0 & 10/10 & 10/10 & 10/10 & 10/10 \\
        1 & 10/10 & 10/10 & 10/10 & 10/10 \\
        2 & 10/10 & 10/10 & 10/10 & 10/10 \\
        3 & 10/10 & 10/10 & 10/10 & 10/10 \\
        4 & 8/10  & 9/10  & 9/10  & 9/10 \\
        5 & 9/10  & 10/10 & 9/10  & 10/10 \\
        6 & 5/10  & 3/10  & 6/10  & 5/10 \\
        7 & 8/10  & 5/10  & 6/10  & \textbf{10/10} \\
        8 & 3/10  & 8/10  & 8/10  & 9/10 \\
        9 & 9/10  & 10/10 & 9/10  & 10/10 \\
        \midrule
        \textbf{Total} & \textbf{82/100} & \textbf{85/100} & \textbf{87/100} & \textbf{93/100} \\
        \bottomrule
    \end{tabular}
\end{table}

\IfFileExists{vvlaps_tsne_plot.png}{%
\section{t-SNE projection of Octo readouts}
\label{app:tsne}

Figure~\ref{fig:tsne} shows a t-SNE projection of the Octo last-layer readouts on libero\_object, coloured by the Monte Carlo value target $G_t$. The visible structure between high- and low-target states is consistent with the SAFE finding of \citet{gu2025safemultitaskfailuredetection} that VLA latent features carry outcome-discriminative signal.

\begin{figure}[!h]
    \centering
    \includegraphics[width=0.45\textwidth]{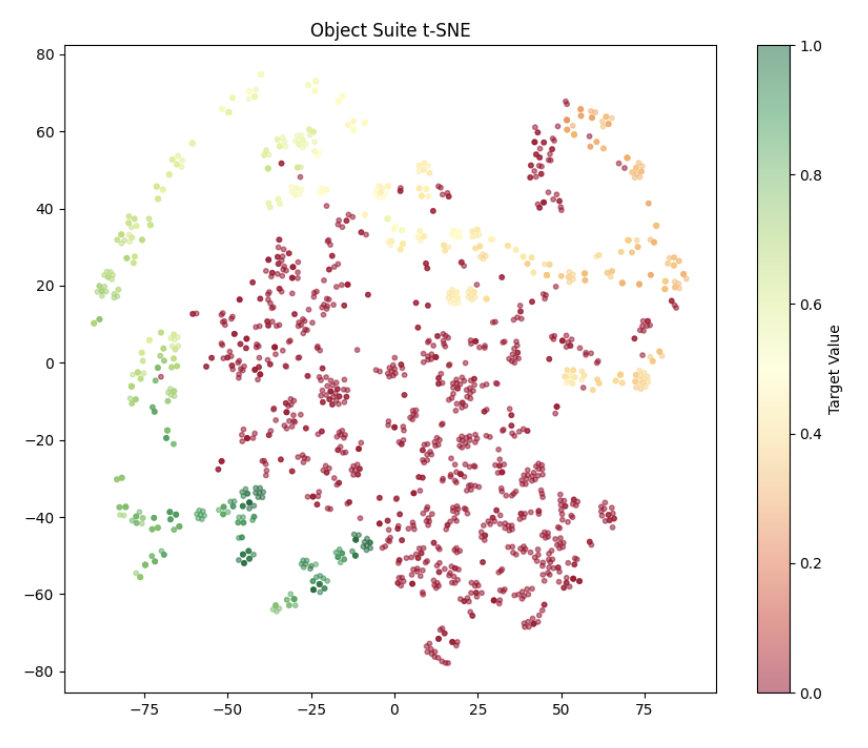}
    \caption{t-SNE projection of Octo last-layer readouts $h_t$ on libero\_object, coloured by Monte Carlo value target $G_t$. Successful and failed rollouts occupy visually distinct regions of the embedding.}
    \label{fig:tsne}
\end{figure}
}{}

\end{document}